\documentclass[10pt,twocolumn,letterpaper]{article}

\usepackage{cvpr}              

\usepackage{graphicx}
\usepackage{amsmath}
\usepackage{amssymb}
\usepackage{booktabs}
\usepackage{multirow}
\usepackage{tabularx}
\usepackage[ruled]{algorithm2e}
\usepackage[accsupp]{axessibility}  
\usepackage[pagebackref,breaklinks,colorlinks]{hyperref}

\usepackage[capitalize]{cleveref}
\crefname{section}{Sec.}{Secs.}
\Crefname{section}{Section}{Sections}
\Crefname{table}{Table}{Tables}
\crefname{table}{Tab.}{Tabs.}

\begin{document}


\title{EVE: Efficient zero-shot text-based Video Editing \\ with Depth Map Guidance and Temporal Consistency Constraints}

\author{Yutao Chen\textsuperscript{\rm 1}{\footnotemark[1]}, \quad Xingning Dong\textsuperscript{\rm 2}{\footnotemark[1]}, \quad Tian Gan\textsuperscript{\rm 1}
\\
\quad Chunluan Zhou\textsuperscript{\rm 2}, \quad Ming Yang\textsuperscript{\rm 2}, \quad Qingpei Guo\textsuperscript{\rm 2}{\footnotemark[2]}
\\
\normalsize{\textsuperscript{\rm 1}Shandong University, \quad \quad \textsuperscript{\rm 2}Ant Group}
\\
{\tt\small yt-chen@mail.sdu.edu.cn, \quad dongxingning1998@gmail.com, \quad gantian@sdu.edu.cn}
\\
{\tt\small CZHOU002@e.ntu.edu.sg, \quad m-yang4@u.northwestern.edu, \quad qingpei.gqp@antgroup.com}
}


\twocolumn[
{%
\maketitle
\renewcommand\twocolumn[1][]{#1}%
\begin{minipage}{0.98\linewidth}
            \vspace{0.0cm}
		\includegraphics[width=1.0\textwidth]{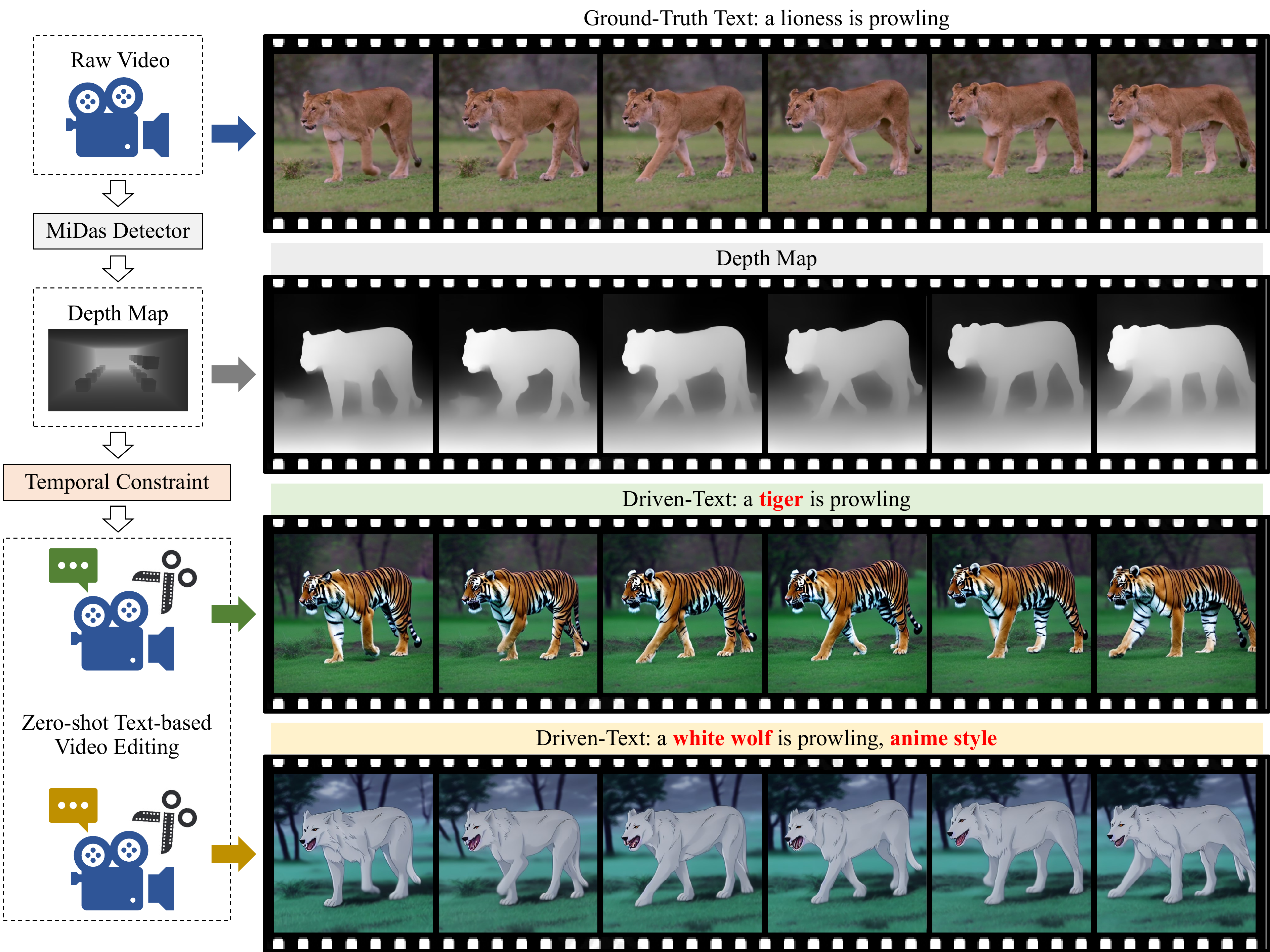}
		\centerline{\footnotesize{We present EVE, an efficient and robust zero-shot text-based video editor, which successfully trades off editing performance and efficiency.}}
		\label{ab-pvt-1}
		\vspace{0.8cm}
\end{minipage}
}]

\renewcommand{\thefootnote}{\fnsymbol{footnote}}
\footnotetext[1]{Yutao Chen and Xingning Dong contribute equally to this manuscript.}
\footnotetext[2]{Qingpei Guo is the corresponding author.}

\begin{abstract}

Motivated by the superior performance of image diffusion models, more and more researchers strive to extend these models to the text-based video editing task. Nevertheless, current video editing tasks mainly suffer from the dilemma between the high fine-tuning cost and the limited generation capacity. Compared with images, we conjecture that videos necessitate more constraints to preserve the temporal consistency during editing. Towards this end, we propose EVE, a robust and efficient zero-shot video editing method. Under the guidance of depth maps and temporal consistency constraints, EVE derives satisfactory video editing results with an affordable computational and time cost. Moreover, recognizing the absence of a publicly available video editing dataset for fair comparisons, we construct a new benchmark ZVE-50 dataset. Through comprehensive experimentation, we validate that EVE could achieve a satisfactory trade-off between performance and efficiency. We will release our dataset and codebase to facilitate future researchers.

\end{abstract}


\section{Introduction}
\label{sec:intro}

Owing to powerful diffusion models \cite{zhang2023text, li2023diffusion}, recent years have witnessed dramatic progress in text-based image synthesis and editing tasks, igniting the soaring research interest in extending these methods to the video editing field. Nevertheless, current text-based video editing methods, which manipulate attributes or styles of videos under the guidance of the driven text, mainly suffer from the dilemma between the considerable fine-tuning cost and the unsatisfied generation performance.

Recent video editing methods could be roughly divided into two classes: tuning-based methods \cite{zhao2023controlvideo, wu2022tune} and zero-shot ones \cite{couairon2023videdit, qi2023fatezero}. The former approaches mainly rely on fine-tuning image diffusion models to derive strong generative priors. Nevertheless, they are usually costly as the fine-tuning step would consume substantial time and GPUs. Towards this end, zero-shot video editing methods aim to directly edit real-world videos without time-consuming fine-tuning. Nevertheless, the edited videos in the zero-shot manner may suffer from the spatio-temporal distortion and inconsistency. Besides, some zero-shot methods are built upon diffusion models fine-tuned on video datasets, which may not be free of the high cost as the tuning-based ones.

In this paper, we attempt to achieve a trade-off between editing performance and efficiency. Specifically, we adopt the approach of zero-shot video editing, while improving editing performance based upon initial image diffusion models rather than video tuning-based ones. Consequently, the primary challenge is how to preserve and improve the temporal consistency of edited videos.

Let's begin by considering human editing. When dealing with images, adjusting object appearances or attributes is relatively straightforward. However, when it comes to videos, a comprehensive evaluation of all edited frames becomes imperative to prevent the spatio-temporal distortion and inconsistency in edited videos. As a result, we conjecture that \textbf{videos necessitate more temporal constraints} to preserve the time consistency, whose editing process could not be as unconstrained as images. This hypothesis also interprets unsatisfied performance when directly extending image diffusion models to videos, as current image editing methods seldom enforce explicit constraints.

Given this argument, different from current methods that neither explicitly control over individual frame editing nor enforce additional constricts on inter-frame generation, we propose two strategies to reinforce temporal consistency constraints during zero-shot video editing: 1) \textbf{Depth Map Guidance}. Depth maps locate spatial layouts and motion trajectories of moving objects, providing robust prior cues for the given video. Therefore, we incorporate depth maps into video editing to improve the temporal consistency. 
And 2) \textbf{Frame-Align Attention}. We enhance the temporal encoding by forcing models to place their attentions on both previous and current frames. 

Moreover, by narrowing the gap of whether introducing depth maps into the noise-to-image inference procedure, we design an efficient parameter optimization strategy that directly updates target latent features without fine-tuning the complex diffusion model. In this way, it takes about 83.1 seconds to edit a video with 8 frames on average.

Currently, there lack public video editing datasets for fair performance comparisons. Towards this end, we construct a new ZVE-50 dataset, where each collected video is associated with four corresponding driven text. We conduct extensive experiments to benchmark our ZVE-50 dataset.

Our contributions are summarized in four folds:

\begin{itemize}

	\item We propose EVE, a zero-shot text-based video editor with a satisfactory trade-off between the generation capability and efficiency.
	
	\item We argue the indispensability of temporal consistency constraints in the video editing task. Towards this end, we propose two strategies to improve the temporal consistency, achieving robust editing performance.

        \item We construct a new benchmark ZVE-50 dataset. To the best of our knowledge, ZVE-50 is the first dataset for zero-shot text-based video editing, which facilitates future researchers to perform a fair comparison.
	
	\item We conduct extensive experiments on ZVE-50 dataset. Experimental results indicate that the proposed EVE is a robust and efficient zero-shot video editing method.

\end{itemize}


\section{Related Work}

\noindent\textbf{Diffusion Models.} Large-scale diffusion models \cite{rombach2022high, croitoru2023diffusion, dhariwal2021diffusion} have achieved start-of-the-art performance in image synthesis and translation. Diffusion models, in essence, are generative probabilistic models that approximate a data distribution $p(x)$ by gradually denoising a normally distributed variable. Nevertheless, training a diffusion model from scratch is often expensive and time-consuming.


\vspace{0.2cm}

\noindent\textbf{Text-based Image Editing.} Text-based image editing \cite{mokady2023null, zhang2023sine, choi2023custom} aims to manipulate the attributes or styles of one image with the guidance of the driven text. Based on powerful diffusion models, researchers have proposed various methods. \textit{E}.\textit{g}., DreamBooth \cite{ruiz2023dreambooth} proposes a subject-driven generation technology by fine-tuning diffusion models, while T2I-Adapters \cite{mou2023t2i} provides an efficient image editing approach with a low training cost.


\vspace{0.2cm}

\noindent\textbf{Text-based Video Generation and Editing.} Motivated by text-based image editing, video editing \cite{molad2023dreamix, lee2023shape} has attracted increasing research interest recently, which could be roughly divided into two categories: tuning-based ones \cite{esser2023structure} and zero-shot ones \cite{qi2023fatezero}. The former approaches mainly edit video attributes by fine-tuning powerful image diffusion models, whose training cost is inevitably expensive. Alternatively, FateZero \cite{qi2023fatezero} proposes the zero-shot video editing task, attempting to generate a text-driven video without extra optimization on complicated generative priors. Nevertheless, FateZero suffers from the limited video editing performance due to weak constraints on the temporal consistency. Moreover, FateZero still heavily relies on Tune-A-Video \cite{wu2022tune}, which is a tuning-based diffusion model and is still costly and time-consuming.


\begin{figure*}[t]
	\centering
	\includegraphics[width=0.98\textwidth]{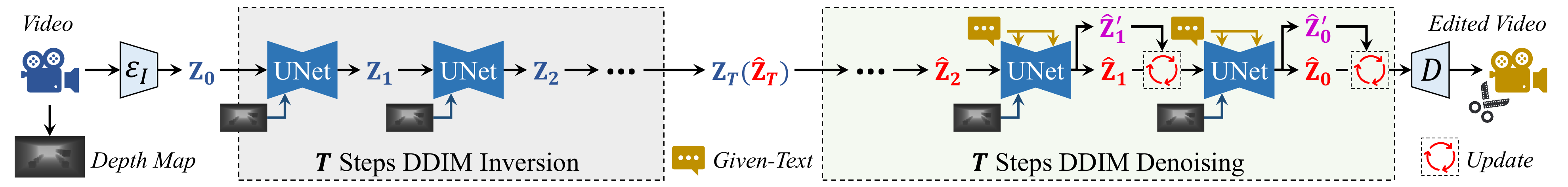}
	\caption{The SIMPLIFIED version of the proposed EVE, presenting the overall video editing pipeline.}
	\label{method_simplied}
        \vspace{0.4cm}
\end{figure*}

\begin{figure*}[t]
	\centering
	\includegraphics[width=0.98\textwidth]{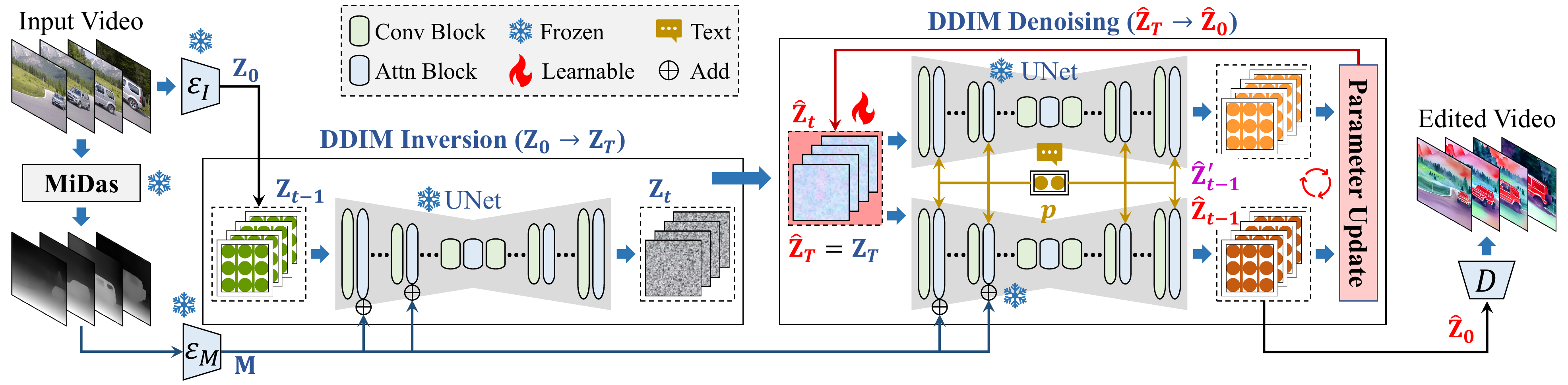}
	\caption{The ELABORATED version of the proposed EVE, detailing the DDIM inversion and denoising procedures.}
	\label{method_complex}
\end{figure*}

\section{Methodology}

\subsection{Preliminary: DMs, LDMs, and DDIMs}

\noindent\textbf{Diffusion Models} (DMs) \cite{sohl2015deep} are essentially generative probabilistic models that approximate a data distribution $p(x)$ by gradually denoising a normally distributed variable. Specifically, diffusion models learn to reconstruct the reverse process of a fixed forward Markov chain ${x}_{1}, {x}_{2}, \cdots, {x}_{T}$, where ${T}$ is the length. The forward Markov chain ($1 \rightarrow T$) could be treated as an image-to-noise procedure, where each Markov transition step $q({x}_{t}|{x}_{t-1})$ is usually formulated as a Gaussian distribution ($\mathcal{N}$) with a variance schedule ${\beta}_{t} \in (0,1)$, that is:
\begin{equation}
\label{forward-SD}
    q({x}_{t}|{x}_{t-1}) = \mathcal{N}({x}_{t}; \sqrt{1-{\beta}_{t}}{x}_{t-1}, {\beta}_{t}{\mathbf{I}}). 
\end{equation}

The reverse Markov chain ($T \rightarrow 1$) could be treated as a noise-to-image procedure, where each reverse Markov transition step $p({x}_{t-1}|{x}_{t})$ is formulated as:
\begin{equation}
\label{backward-SD}
    {p}_{\theta}({x}_{t-1}|{x}_{t}) = \mathcal{N}({x}_{t-1}; {\mu}_{\theta}({x}_{t},t), {\Sigma}_{\theta}({x}_{t},t)),
\end{equation}
where $\theta$ denotes learnable parameters to guarantee that the reverse process is close to the forward one. 

Empirically, current diffusion models could be interpreted as an equally weighted sequence of denoising auto-encoders ${\epsilon}_{\theta}({x}_{t},t)$, which is utilized to recover a denoised variant of their input ${x}_{t}$, and ${x}_{t}$ is a noisy version of the input ${x}$. The optimization objective could be simplified as:
\begin{equation}
\label{optimization-SD}
    \mathbb{E}_{x,\epsilon \sim \mathcal{N}(0,1),t} [{\| \epsilon-{\epsilon}_{\theta}({x}_{t},t) \|}_{2}^{2} ].
\end{equation}

\vspace{0.2cm}

\noindent\textbf{Latent Diffusion Models} (LDMs) \cite{rombach2022high} are trained in the learned latent space ${z}_{t}$ rather than redundant spatial dimensionality ${x}_{t}$, aiming to remove the noise added to latent image features ${\epsilon}_{x}$. LDMs are generally composed of an encoder $\mathcal{E}$, a time-conditional UNet $\mathcal{U}$, and a decoder $\mathcal{D}$, where $z=\mathcal{E}(x)$ and $x \approx \mathcal{D}(\mathcal{E}(x))$. The optimization objective could be formulated as:
\begin{equation}
\label{optimization-LSD}
    \mathbb{E}_{{\epsilon}_{x},\epsilon \sim \mathcal{N}(0,1),t} [{\| \epsilon-{\epsilon}_{\theta}({z}_{t},t) \|}_{2}^{2} ].
\end{equation}

\vspace{0.2cm}

\noindent\textbf{Denoising Diffusion Implicit Models} (DDIMs) \cite{mokady2023null} could accelerate the sampling from the distribution of images/videos at the denoising step. During inference, deterministic DDIM sampling ($T \rightarrow 1$) aims to recover a clean latent ${z}_{0}$ from a random noise ${z}_{T}$ with a noise schedule ${\alpha}_{t}$, which could be formulated as: 
\begin{equation}
\label{DDIM-to}
    {z}_{t-1} = \sqrt{ \frac{{\alpha}_{t-1}}{{\alpha}_{t}}}{z}_{t} + (\sqrt{1-{{\alpha}_{t-1}}} - \sqrt{\frac{1}{{\alpha}_{t}}-1}) \cdot {\epsilon}_{\theta}.
\end{equation}

On the contrary, DDIM inversion ($1 \rightarrow T$) aims to process a clean latent ${z}_{0}$ into a noise one ${\hat{z}}_{T}$, which could be simplified as: 
\begin{equation}
\label{DDIM-from}
    {\hat{z}}_{t} = \sqrt{ \frac{{\alpha}_{t}}{{\alpha}_{t-1}}}{\hat{z}}_{t-1} + (\sqrt{1-{{\alpha}_{t}}} - \sqrt{\frac{1}{{\alpha}_{t-1}}-1}) \cdot {\epsilon}_{\theta}.
\end{equation}

Compared with conventional DMs that directly employ random noise as inputs and attempt to map each noise vector to a specific image, we exploit DDIM inversion to produce a $T$ steps trajectory between the clean latent ${z}_{0}$ to a Gaussian noise vector ${z}_{T}$. Then we treat ${z}_{T}$ as the start vector of the denoising step. This configuration seems appropriate for our video editing task, since it ensures that the generated video would be close to the original one.

Note that we employ \textbf{LDMs} and \textbf{DDIM inversion/denoising} in zero-shot text-based video editing. Readers can refer to \cite{rombach2022high} (LDM) and \cite{song2020denoising} (DDIMs) for more details of formulation derivations if necessary.

\subsection{Problem Formulation}

Given a video $V$ and a prompt text $P$, zero-shot text-based video editing aims to generate an edited video $\hat{V}$, which aligns with the description outlined in the prompt $P$ and looks similar to the original video $V$.

\subsection{Overall Framework}

As shown in Figure \ref{method_simplied} and \ref{method_complex}, we present both simplified and elaborated versions of the overall framework. The simplified version could be treated as a flow chart that reveals the whole processing pipeline of our EVE. While the complex one presents detailed information mainly on the iterative DDIM inversion and denoising procedures. 

As shown in Figure \ref{method_simplied}, our EVE is built upon the pre-trained latent diffusion model (LDM), which is composed of a UNet for T-timestep DDIM inversion and denoising. To enforce the temporal consistency of the generated video, we introduce depth maps and exploit them to guide the editing process. Moreover, we propose two consistency constraints to prevent edited videos from spatial or temporal distortion. 

We first present the overall pipeline of our EVE based upon Figure \ref{method_complex}, including the following five steps.

\vspace{0.2cm}

\noindent\textbf{1. Frozen Features Extraction}. Given a video $V$, we first derive $K$ frames from $V$, and utilize an image encoder $\mathcal{E}_{I}$ to obtain \textbf{frozen} latent features ${\mathbf{Z}}_{0} = \mathcal{E}(V)$, where ${\mathbf{Z}}_{0} = \{{z}_{0}^{i}\}_{i=1}^{K}$. Meanwhile, we employ the MiDas Detector \cite{ranftl2020towards} to generate $K$ depth maps from $V$, and utilize another visual encoder $\mathcal{E}_{M}$ to obtain \textbf{frozen} depth-map features ${\mathbf{M}} = \{{m}^{i}\}_{i=1}^{K}$. Moreover, we utilize a text encoder $\mathcal{E}_{p}$ to process the prompt $P$ into \textbf{frozen} features $p$.

\vspace{0.2cm}

\noindent\textbf{2. DDIM Inversion}. Then we repeat DDIM inversion for $T$ steps to derive Gaussian noise vectors ${\textbf{Z}}_{T}$ from video latent features ${\textbf{Z}}_{0}$. Each DDIM inversion at timestep $t$ could be formulated as:
\begin{equation}
\label{DDIM-INV}
    {\mathbf{Z}}_{t} = {\rm {DDIM}_{inv}}({\mathbf{Z}}_{t-1} \ | \ {\mathbf{M}},t) \quad t=1 \rightarrow T,
\end{equation}
where ${\rm {DDIM}_{inv}}$ denotes DDIM inversion shown in Eq. \ref{DDIM-from}. 

To prevent the edited video from temporal distortion and inconsistency, we improve the image-based DDIM inversion operation by introducing depth-map features into the down-sampling pass of the \textbf{frozen} UNet, which could rectify the discrepancies among neighboring frames at each inversion step. In this way, we ensure that the generated noise vectors ${\textbf{Z}}_{T}$ would not severely spoil the temporal consistency.


Specifically, we repeat $T$ DDIM inversion steps to process video latent features ${\mathbf{Z}}_{0}$ into generated noise vectors ${\mathbf{Z}}_{T}$

\vspace{0.2cm}

\noindent\textbf{3. DDIM Denoising}. Afterward, we repeat DDIM denoising for $T$ steps to obtain edited video features ${\mathbf{\hat{Z}}}_{0}$ from DDIM inverted noise ${\mathbf{\hat{Z}}}_{T}$, where ${\mathbf{\hat{Z}}}_{T} = {\mathbf{Z}}_{T}$. Each DDIM denoising at timestep $t$ could be formulated as:
\begin{equation}
\label{DDIM-DEN}
    {\mathbf{\hat{Z}}}_{t-1} = {\rm {DDIM}_{den}}({\mathbf{\hat{Z}}}_{t} \ | \ p, {\mathbf{M}, t}), \quad t=T \rightarrow 1,
\end{equation}
where ${\rm {DDIM}_{den}}$ denotes DDIM denoising shown in Eq. \ref{DDIM-to}. 

To prevent the edited video from temporal distortion and inconsistency, we improve the image-based DDIM denoising operation from two aspects: 1) We introduce depth-map features into the down-sampling pass of the \textbf{frozen} UNet as DDIM inversion. And 2) we propose the frame-aligned attention to place explicit temporal constraints on the edited video, which is discussed in the following subsection.

Specifically, we repeat $T$ DDIM denoising steps to otbain edited video features ${\mathbf{\hat{Z}}}_{0}$ from DDIM inverted noise ${\mathbf{\hat{Z}}}_{T}$

\vspace{0.2cm}

\noindent\textbf{4. Parameter Optimization}. To reduce the computation cost and make the generation process more efficient, we freeze all feature extractors (\textit{i}.\textit{e}., $\mathcal{E}_{I}$, $\mathcal{E}_{M}$, and $\mathcal{E}_{P}$) and Unets, and only set noise vectors ${\mathbf{\hat{Z}}}_{t}$ in DDIM denoising to be trainable. In another word, different from conventional editing methods that update ``neural networks", we directly update ``latent noise" to obtain edited videos. 

Specifically, at each timestep $t$ in DDIM denoising, except for ${\mathbf{\hat{Z}}}_{t-1}$, we also derive auxiliary vectors ${\mathbf{\hat{Z}}}_{t-1}^{'}$ as:
\begin{equation}
\label{DDIM-AUX}
    {\mathbf{\hat{Z}}}_{t-1}^{'} = {\rm {DDIM}_{den}}({\mathbf{\hat{Z}}}_{t} \ | \ p, t), \quad t=T \rightarrow 1.
\end{equation}

Compared with ${\mathbf{\hat{Z}}}_{t-1}$ (Eq. \ref{DDIM-DEN}), ${\mathbf{\hat{Z}}}_{t-1}^{'}$ is obtained without strict depth map constraints, which could be treated as free image editing that could unleash the generation capacity of powerful image-based diffusion models. In brief, ${\mathbf{\hat{Z}}}_{t-1}$ sacrifices the creativity to preserve the temporal consistency, while ${\mathbf{\hat{Z}}}_{t-1}^{'}$ is just the opposite. Therefore, we leverage the more creative ${\mathbf{\hat{Z}}}_{t-1}^{'}$ and more temporal consistent ${\mathbf{\hat{Z}}}_{t-1}$, pursuing to achieve a trade-off between diversity and quality.

The detailed DDIM denoising procedure is illustrated in Algorithm \ref{alg_update}, including the parameter optimization step (Lines 4-5). ${\Delta}_{x}(\mathcal{L})$ denotes updating trainable $x$ by the gradient descent procedure according to the loss $\mathcal{L}$, and ${\rm cosin}(\cdot,\cdot)$ denotes the cosine similarity computation.

\begin{algorithm}
	\caption{DDIM Denoising Procedure.}
	\label{alg_update}
	\LinesNumbered
	\KwIn{DDIM inverted noise ${\mathbf{\hat{Z}}}_{T}$, text prompt features $p$, depth-map features $\mathbf{M}$, learning rate $\lambda$}

	\KwOut{edited video features ${\mathbf{\hat{Z}}}_{0}$} 

        \vspace{0.2cm}
	
	\For{$i \leftarrow T$ \KwTo $1$}{
	
	    ${\mathbf{\hat{Z}}}_{t-1} = {\rm {DDIM}_{den}}({\mathbf{\hat{Z}}}_{t} \ | \ p, {\mathbf{M}},t)$ \;
	    
	    ${\mathbf{\hat{Z}}}_{t-1}^{'} = {\rm {DDIM}_{den}}({\mathbf{\hat{Z}}}_{t} \ | \ p,t)$    \;

            $\mathcal{L} = 1 - {\rm cosin}({\mathbf{\hat{Z}}}_{t-1}, \mathbf{\hat{Z}}_{t-1}^{'})$

            $\mathbf{\hat{Z}}_{t-1} = \mathbf{\hat{Z}}_{t-1} - \lambda {\Delta}_{{\hat{Z}}_{t-1}}(\mathcal{L})$

	}
	
\vspace{-0.1cm}
\end{algorithm}

\noindent\textbf{5. Edited Video Decoding}. Ultimately, we feed the \textbf{frozen} visual decoder $\mathcal{D}$ with generated latent features ${\mathbf{\hat{Z}}}_{0}$, obtaining the edited video $\hat{V}=\mathcal{D}({\mathbf{\hat{Z}}}_{0})$.

\subsection{Temporal Consistency Constraints}

As aforementioned, we assume that videos necessitate more temporal constraints to preserve the time consistency. Therefore, we propose two strategies to alleviate temporal distortion and inconsistency problems.

\vspace{0.2cm}

\noindent\textbf{1. Depth Map Guidance}. Depth maps record visual representations of the distance information, revealing spatial layouts and motion trails of all objects within a video. Therefore, depth maps could be treated as strong prior cues to guide the video editing procedure close to the initial version. Nevertheless, recent video editing methods seldom take advantage of depth maps and neglect the significance of explicitly intervening in the video editing procedure, resulting in intractable temporal distortion and inconsistency problems. Towards this end, we introduce depth maps into the down-sampling pass of the frozen UNet for both DDIM inversion and denoising procedures, forcing the editing process to imitate motion trails and scene transformations of the origin video. In this way, the stability and consistency of the edited video would be improved.



\vspace{0.2cm}

\begin{figure}[t]
	\centering
	\includegraphics[width=0.49\textwidth]{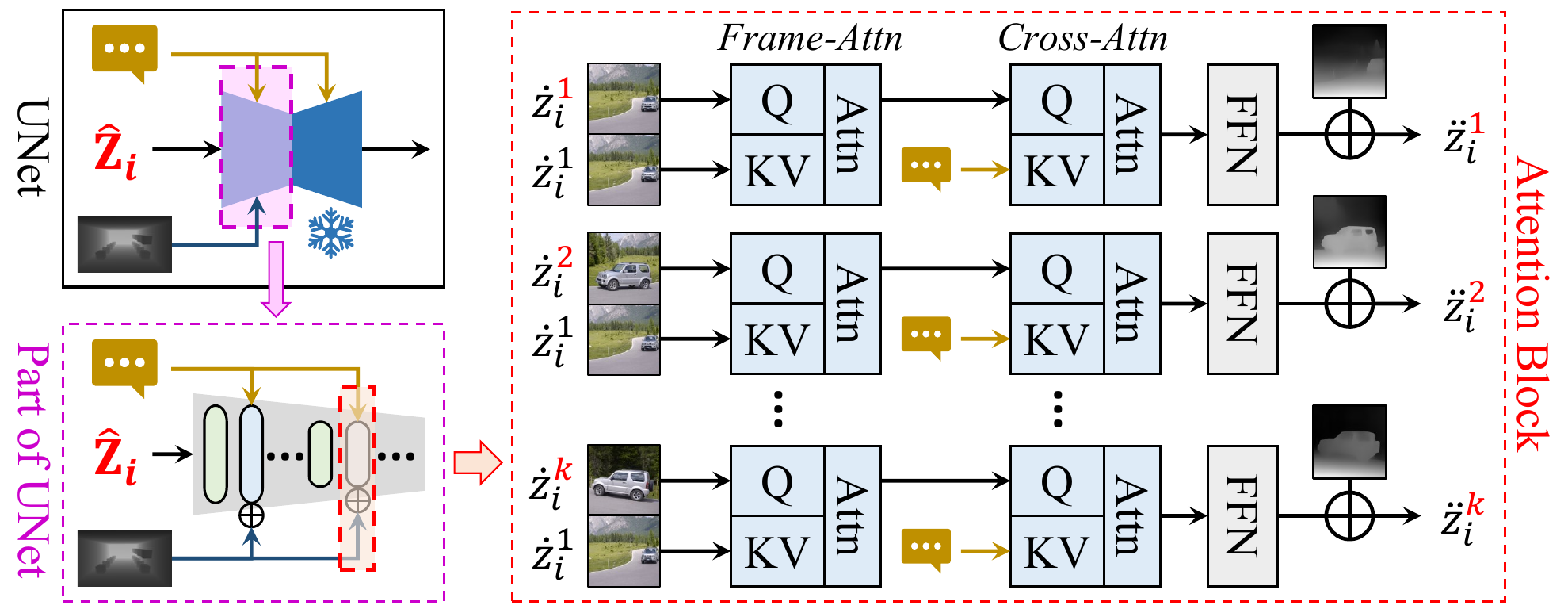}
	\caption{The architecture of the attention block within the UNet. Note that we propose the Frame-Align Attention to improve the temporal consistency.}
	\label{method_attn}
\end{figure}

\noindent\textbf{2. Frame-Align Attention}. We propose the frame-align attention (FAA) to explicitly introduce the temporal information during video editing. As illustrated in Figure \ref{method_attn}, a typical UNet comprises a series of ``Conv-Attn" blocks to conduct the down-sampling and up-sampling calculation. The conventional attention block (Attn) contains a self-attention (SA) module \cite{vaswani2017attention}, a cross-attention (CA) module \cite{zheng2021deep}, and a feed-forward network (FFN). The computation of ${\rm SA}(Q,K,V)$ and ${\rm CA}(Q,K,V)$ could be formulated as: 
\begin{equation}
    \label{QKV-old}
    \left \{
    \begin{aligned}
	{\rm{SA}}: Q={W}^{Q}{\dot{z}}^{i}, K={W}^{K}{\dot{z}}^{i}, V={W}^{V}{\dot{z}}^{i},
	\\
	{\rm{CA}}: Q={W}^{Q}{\dot{z}}^{i}, K={W}^{K}{p}, \ \; V={W}^{V}{p},
	\end{aligned}
	\right.
\end{equation}
where ${W}$ denotes \textbf{frozen} projection matrices, ${\dot{z}}^{i}$ is the latent features of the ${i}^{th}$ frame within the video, and ${p}$ is the latent features of the text prompt.

Conventional self-attention modules are inherited from image diffusion models, which encodes each frame separately and seem insufficient in preserving the temporal consistency for video editing. Therefore, we propose the frame-align attention (FAA) to replace $K$ and $V$ with the first frame features ${\dot{z}}^{1}$, forcing models to emphasize both previous and current frames for better temporal encoding. The computation of ${\rm FAA}(Q,K,V)$ could be formulated as:
\begin{equation}
\label{QKV-FAA}
    {\rm{FAA}}: Q={W}^{Q}{\dot{z}}^{i}, K={W}^{K}{\dot{z}}^{1}, V={W}^{V}{\dot{z}}^{1}.
\end{equation}




\begin{table*}[t]
	\small
	\vspace{-0.4cm}
	\centering
	\begin{tabular}{p{0.6cm}<{\centering}|p{1.5cm}<{\centering}|p{1.0cm}<{\centering}|p{1.0cm}<{\centering}
	|p{0.7cm}<{\centering}p{0.7cm}<{\centering}p{0.7cm}<{\centering}p{0.7cm}<{\centering}p{0.7cm}<{\centering}
	|p{0.7cm}<{\centering}p{0.7cm}<{\centering}p{0.7cm}<{\centering}p{0.7cm}<{\centering}p{0.7cm}<{\centering}}
		\hline
            \multicolumn{1}{c|}{\multirow{2}{*}{No.}} &
		\multicolumn{1}{c|}{\multirow{2}{*}{Model}} &
		\multicolumn{1}{c|}{\multirow{2}{*}{DMG}} &
            \multicolumn{1}{c|}{\multirow{2}{*}{Attn}} &
		\multicolumn{5}{c|}{Temporal Consistency} &
		\multicolumn{5}{c}{Prompt Consistency}
		\\ \cline{5-14} 
            \multicolumn{1}{c|}{} &
		\multicolumn{1}{c|}{} &
		\multicolumn{1}{c|}{} &
		\multicolumn{1}{c|}{} &
		\multicolumn{1}{c}{OR} &
            \multicolumn{1}{c}{OA} &
            \multicolumn{1}{c}{ST} &
            \multicolumn{1}{c}{BC} &
            \multicolumn{1}{c|}{AVG} &
		\multicolumn{1}{c}{OR} &
            \multicolumn{1}{c}{OA} &
            \multicolumn{1}{c}{ST} &
		\multicolumn{1}{c}{BC} &
		\multicolumn{1}{c}{AVG}
		\\ \hline
		\multicolumn{14}{c}{\multirow{1.5}{*}{ \textit{Performance comparisons between the proposed EVE and FateZero. }}}
		\vspace{0.5em} \\\hline
		
		A1 & FateZero & -  & STSA & 95.53 & 95.64 & 95.84 & 96.62 & \underline{95.91} & 28.92 & 29.43 & 29.24 & 28.90 & \underline{29.12} \\
		A2 & EVE & $\surd$ & FAA & \textbf{96.41} & \textbf{96.65} & \textbf{96.43} & \textbf{96.70} & \underline{\textbf{96.54}} & \textbf{30.39} & \textbf{31.01} & \textbf{31.27} & \textbf{28.94} & \underline{\textbf{30.40}} \\       \hline
		\multicolumn{14}{c}{\multirow{1.5}{*}{ \textit{Ablation study of two proposed temporal consistency constraints within EVE.}}}
		\vspace{0.5em} \\\hline
            B1 & & -  & SA & 92.59 & 92.94 & 92.07 & 96.30 & \underline{93.48} & 25.39 & 26.43 & 27.38 & 28.88 & \underline{27.02} \\
            B2 &  & - & FAA & 95.26 & 95.20 & 95.16 & 95.68 & \underline{95.33} & 27.77 & 29.37 & 29.11 & 29.71 & \underline{28.99} \\ 
            B3 & EVE & $\surd$  & SA & 94.57 & 94.02 & 94.88 & 95.45 & \underline{94.73} & \textbf{30.58} & \textbf{31.12} & 31.16 & 28.45 & \underline{30.33} \\ 
    	
            B4  & & $\surd$  & SCA & 95.74 & 96.18 & 96.13 & 96.62 & \underline{96.17} & 30.26 & 31.01 & 31.00 & 28.91 & \underline{30.29} \\
            B5  & & $\surd$  & FAA & \textbf{96.41} & \textbf{96.65} & \textbf{96.43} & \textbf{96.70} & \underline{\textbf{96.54}} & 30.39 & 31.01 & \textbf{31.27} & \textbf{28.94} & \underline{\textbf{30.40}} \\ 
		
		 \hline
		
	\end{tabular}
\vspace{0.0cm}
\caption{Performance comparisons between our EVE with FateZero, and ablation study of two proposed temporal consistency constraints within EVE. We report the detailed results of four video editing missions and their average performance (\underline{underlined}), where \textit{OR = Object Replacement}, \textit{OA = Object Adding}, \textit{ST = Style Transfer}, and \textit{BC = Background Changing}. All experiments are conducted on one A40 GPU under the same setting. ``DMG" denotes with/without the depth map guidance.}
\label{ablation_study}
\end{table*}

\section{Experiments}

\subsection{Dataset Construction}

Since zero-shot text-based video editing is a novel task, to the best of our knowledge, there lacks a public dataset to perform fair performance and efficiency comparisons. Towards this end, we construct \textit{Zero-shot Video Editing 50} (dubbed as \textbf{ZVE-50}) to fulfill this job.

\vspace{0.2cm}

\noindent\textbf{Data Collection}. We collect videos from two resources: DAVIS-2017 \cite{pont20172017} and stock-video-footage \footnote{https://www.videvo.net/stock-video-footage/}. DAVIS-2017 is a competition dataset for the video object segmentation task \cite{yao2020video}, while stock-video-footage is a public website for free stock video clips and motion graphics. After filtering out videos with similar scenes and styles to avoid the repeatability and promote the diversity, we collect 14 short videos from DAVIS-2017 and 36 ones from stock-video-footage, resulting in the ZVE-50 dataset.

\vspace{0.2cm}

\noindent\textbf{Caption Generation}. Then we feed collected videos into BLIP2 \cite{li2023blip} to obtain the corresponding captions. Specifically, we generate several candidate captions and select the longest one as the ground-truth text. 

\vspace{0.2cm}

\noindent\textbf{Prompt Generation}. Afterward, we employ GPT-4 \footnote{https://openai.com/gpt-4} to generate the driven text derived from video captions and our manually made prompts. There are four types of driven text, requiring models to edit the given video by 1) Object Replacement (OR), 2) Object Adding (OA), 3) Style Transfer (ST), and 4) Background Changing (BC). Here we present an example of feeding GPT-4 with the manually written prompt and ground-truth caption to obtain the driven text: 

Q (human): \textit{Here is a sentence. Please replace the object with another object with a similar shape: ``a pink lotus flower in the water with green leaves"}

A (GPT-4): \textit{a pink lotus flower floating in a tranquil koi pond with lily pads}

Ultimately, we manually check all videos, captions, and prompt text to ensure the correctness of the ZVE-50 dataset.

\subsection{Experimental Settings}

\noindent\textbf{Implementation Details}. Zero-shot text-based video editing directly takes a given video and outputs its edited version, which differs from previous methods with explicit training or testing procedure. Specifically, we freeze the pre-trained Latent Diffusion Model as our basic model, where the visual encoder $\mathcal{E}_{I}$, the UNet, and the visual decoder $\mathcal{D}$ are inherited from \cite{rombach2022high} with the version of v1.5. We employ MiDas \cite{ranftl2020towards} to derive depth maps, and utilize frozen Resnet blocks \cite{he2016deep} to extract depth map features $M$. The text encoder $\mathcal{E}_{p}$ is the pre-trained CLIP text encoder \cite{radford2021learning}. 

During video editing, following \cite{wu2022tune} and \cite{qi2023fatezero}, we uniformly sample 8 frames at the resolution of 512*512 from each video, and conduct DDIM inversion and denoising steps 50 ($T$) times. The learning rate $\lambda$ is 0.8. It takes about \textbf{83 seconds} to edit a video on an A40 GPU.

\noindent\textbf{Evaluation Metrics.}

We employ two metrics, \textit{i}.\textit{e}., Temporal Consistency (TC) and Prompt Consistency (PC), to thoroughly evaluate the quality of edited videos: 1) Following \cite{esser2023structure}, we first extract CLIP embedding of all frames within the edited video, and calculate the average cosine similarity between all pairs of neighborhood frames to derive the Temporal Consistency score. 2) Following \cite{qi2023fatezero}, we utilize Text-Video CLIP Score to evaluate the Prompt Consistency between the edited video $\hat{V}$ ($K$ frames) and the driven text $p$, which could be formulated as:
\begin{equation}
\label{metric-PC}
    {\rm CLIP}(\hat{V}, p) = \frac{1}{K} {\sum}_{k=1}^{K} {\rm CLIP}(\hat{v}^{k}, p).
\end{equation}


\begin{table}[t]
	\small
	\begin{tabular}{p{1.2cm}<{\centering}|p{1.8cm}<{\centering}|p{1.8cm}<{\centering}|p{1.8cm}<{\centering}}
		\hline
		\multicolumn{1}{c|}{Model} & \multicolumn{1}{c|}{Tune-A-Video}& \multicolumn{1}{c|}{FateZero}&\multicolumn{1}{c}{EVE}
		\\ \hline
		
		Time & $\sim$ 30 minutes &  247.6 seconds &  \textbf{83.1} seconds  \\
		
		\hline
	\end{tabular}
\caption{Efficiency comparisons between our EVE with tuning-based Tune-A-Video and zero-shot FateZero. All experiments are conducted on one A40 GPU.}
\vspace{0.0cm}
\label{comp_eff}
\end{table}


\subsection{Performance and Efficiency Comparisons}

As aforementioned, zero-shot text-based video editing is a novel task without public datasets and widely-employed baselines. 
Thus, it is intractable to conduct a fair performance comparison with other methods. Therefore, we compare the video editing efficiency between our EVE with the tuning-based Tune-A-Video \cite{wu2022tune} and the zero-shot Fatezero \cite{qi2023fatezero}. We also compare the zero-shot video editing performance between our EVE and Fatezero in two quantitative metrics.

Regarding efficiency comparisons, as illustrated in Table \ref{comp_eff}, compared with the tunning-based Tune-A-Video that takes about 30 minutes to generate an edited video, zero-shot video editing methods are much more efficient as they shorten the time to less than 5 minutes. Moreover, compared with the baseline FateZero, the proposed EVE only costs about 1/3 of the total time (83s \textit{vs}. 247s) to edit a video, which is more time-efficient and user-friendly.


Regarding performance comparisons, as illustrated in Table \ref{ablation_study} (A1 \textit{vs}. A2), we observe that our proposed EVE outperforms the baseline FateZero in all four tasks on the constructed ZVE-50 dataset, achieving an average improvement of +0.63$\%$ on the temporal consistency and +1.28$\%$ on the prompt consistency. It indicates that EVE is an efficient and robust video editing method, which improves the temporal consistency of the generated video.


\subsection{Ablation Study}

Based on the argument that video editing necessitates more temporal constraints to preserve the time consistency, we propose two constraints to alleviate temporal distortion and inconsistency problems. We conduct several ablation study to verify their effectiveness on our ZVE-50 dataset.

As illustrated in Table \ref{ablation_study}, we have three observations:

1) Depth maps are strong generative priors that prevent the edited video from temporal distortion and inconsistency. Compared with B2 (without DMG) and B5, we witness an obvious performance decay on both temporal and prompt consistency, indicating the indispensability of the proposed depth map guidance strategy.

2) The proposed Frame-Align Attention (FAA) reinforce the temporal encoding to improve the consistency of edited videos. Compared with B3 (without FAA) and B5, methods equipped with FAA would outperform conventional ones with SA by a large margin, especially on the metric of the temporal consistency.

3) We also compare our FAA with the Sparse-Causal Attention (SCA) mechanism proposed by Tune-A-Video. SCA calculates attentions among current frames and the previous neighborhood ones, which could be formulated as:
\begin{equation}
\label{QKV-SCA}
    {\rm{SCA}}: Q={W}^{Q}{\dot{z}}^{i}, K={W}^{K}[{\dot{z}}^{1};\dot{z}^{i-1}], V={W}^{V}[{\dot{z}}^{1};\dot{z}^{i-1}],
\end{equation}
where $[\cdot]$ denotes the concatenation operation. We implement SCA based upon our backbone with the same setting. Compared with B4 (with SCA) and B5, we outperform SCA on both temporal and prompt consistency in all four tasks, proving the advantages of our proposed FAA strategy.

\subsection{Visualization Results and Applications}

As illustrated in Figure \ref{app_visualization}, our EVE supports four types of applications towards zero-shot text-based video editing: 

1) Object Replacement (OR). OR replaces an object with another one in the given video. \textit{E}.\textit{g}., ``\textit{man} $\rightarrow$ \textit{woman}".

2) Object Adding (OA). OA adds a new object to the original video. \textit{E}.\textit{g}., ``\textit{man} $\rightarrow$ \textit{man with glasses}".

3) Style Transfer (ST). ST transfers the original video into different styles. \textit{E}.\textit{g}., ``\textit{style} $\rightarrow$ \textit{Van Gogh style}".

4) Background Changing (BC). BC changes the video background. \textit{E}.\textit{g}., ``\textit{background} $\rightarrow$ \textit{under stars}".





\section{Conclusion}

We present EVE, a robust and efficient zero-shot text-based video editing method, to tackle with the dilemma between the considerable fine-tuning cost and the unsatisfied generation performance. Motivated by the observation that videos necessitate more constraints to preserve the time consistency, we introduce depth maps and two temporal consistency constraints to guide the video editing procedure. In this way, the proposed EVE achieves a satisfactory trade-off between performance and efficiency. Moreover, we construct and benchmark ZVE-50, a public video editing dataset that provides a fair comparison for future researchers. 


\section{Future Work}
In the future, we aim to further improve the quality of edited videos, narrowing the performance gap between tuning-based video editing methods and zero-shot ones. \textit{E}.\textit{g}., introducing the triplet attention mechanism \cite{zhou2021triplet} into the attention block of Unets to promote the temporal stability; and generating pseudo labels by recording attention maps of Unets during the DDIM inversion procedure, which helps to build a knowledge distillation mechanism \cite{hinton2015distilling} in the following denoising step.

\twocolumn[{%
\renewcommand\twocolumn[1][]{#1}%

\begin{minipage}{0.98\linewidth}
            \vspace{-0.6cm}
		\includegraphics[width=0.98\textwidth]{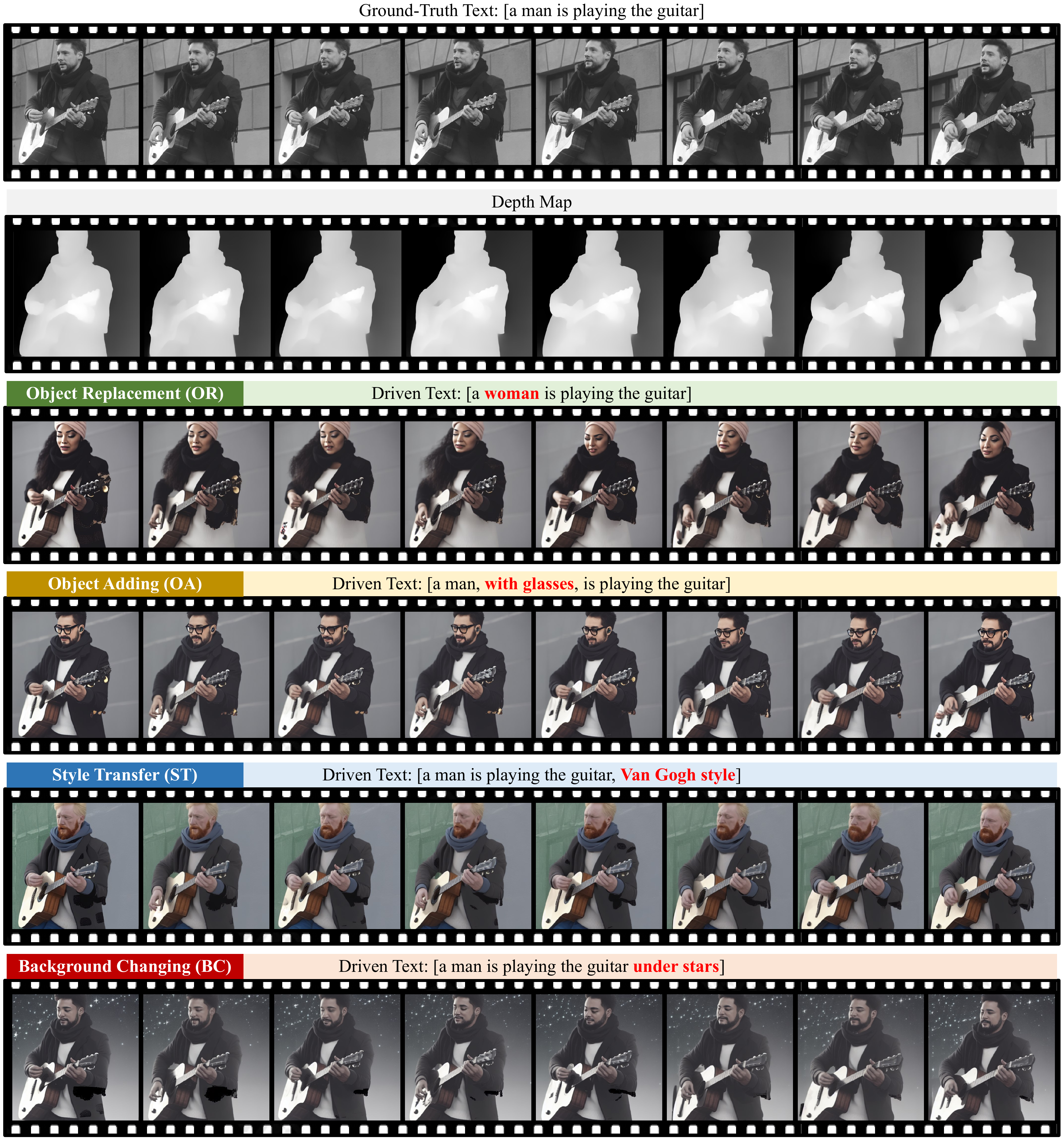}
		\centerline{\footnotesize{Visualization results of the proposed EVE on four applications: Object Replacement, Object Adding, Style Transfer, and Background Changing.}}
		\label{app_visualization}
		\vspace{0.4cm}
\end{minipage}
}]

\newpage
{\small
\bibliographystyle{ieee_fullname}
\bibliography{egbib}
}

\end{document}